\documentclass[runningheads]{llncs}
\usepackage[T1]{fontenc}
\usepackage{graphicx,verbatim}
\usepackage{enumitem}
\usepackage{color}
\usepackage{url}

\usepackage{hyperref}
\usepackage{amsmath}
\usepackage{xcolor}
\usepackage{amssymb}

\usepackage{tikz}
\usepackage{amsmath}
\usetikzlibrary{positioning, arrows.meta, shapes.geometric, calc, fit, backgrounds, decorations.pathreplacing}

\definecolor{dataColor}{RGB}{39, 174, 96}
\definecolor{sslColor}{RGB}{41, 128, 185}
\definecolor{unetColor}{RGB}{142, 68, 173}
\definecolor{adaptColor}{RGB}{243, 156, 18}
\definecolor{boxBg}{RGB}{248, 249, 250}

\newcommand{\virg}[1]{``#1''}

\begin{document}

\title{Geometric-to-Semantic Spherical Transfer Learning for  Cortical Sulci Labeling }
%
\author{Saeb Tounsi\inst{1} \and
Joël Chavas \inst{1} \and
Pietro Gori\inst{2} \and Vincent Frouin \inst{1} \and Denis Rivière \inst{1} \and  Jean-François Mangin \inst{1} }
\authorrunning{S. Tounsi et al.}
%
\institute{ Paris-Saclay University, CEA, NeuroSpin, Baobab, Saclay, France \and
LTCI, Télécom Paris, Institut Polytechnique de Paris, France
\email{tounsi.saeb@gmail.com}\\
 }

\maketitle              
\begin{abstract}
Deep learning on cortical surfaces faces a dilemma: capturing the complex topology of over 60 nomenclature-dependent sulci per hemisphere requires high-capacity models, yet the extreme scarcity of expert annotations ($N=62$ subjects) inevitably causes overfitting. 
Standard supervised approaches fail to generalize on this data-scarce regime, particularly for variable and small sulci where topological ambiguity is high. 
To overcome this limitation, we introduce a Geometric-to-Semantic Spherical Transfer Learning framework. 
First, we leverage massive unlabeled data (UK Biobank with $\approx 30,000$ subjects) to pre-train a spherical encoder using a locally-optimized strategy. 
By relying solely on continuous surface features (curvature and depth), the relevance of this pre-training is confirmed by the model's ability to detect  localized and rare topological traits, such as sulcal interruptions. 
The downstream labeling task, however, introduces extracted sulcal fundi (lines) as an explicit semantic input. To bridge this dimensional domain gap (from purely geometric to semantic) without causing catastrophic forgetting, these anatomical lines are integrated into the pre-trained backbone via a  soft-initialized Topological Prior Injector.
Our experiments demonstrate that this approach outperforms fully supervised baselines trained from scratch, achieving a mean Dice of 0.77. Crucially, a local  analysis reveals that the self-supervised geometric priors yield the largest performance gains on variable and tertiary sulci (up to 14.8\%), confirming that learning the cortex shape  is highly beneficial for identifying its rarest parts.
Code is publicly available at \url{https://anonymous.4open.science/r/2026_slabelin2-72BC}

\end{abstract}

\section{Introduction}

The human cerebral cortex is characterized by a complex pattern of convolutions, where sulci serve as essential landmarks for structural and functional brain mapping. Accurate identification of these folds is crucial for understanding neurodevelopmental trajectories and improving the localization of functional areas. However, cortical sulci exhibit extreme inter-individual variability in terms of morphology, spatial location, length, and depth.

Historically, sulcal labeling has served two primary objectives: driving inter-subject brain alignment and enabling detailed morphometric studies. While the former can tolerate some geometric approximation, the latter demands high topological precision. The automated recognition of cortical sulci relied on volumetric approaches. Pioneering frameworks, such as BrainVISA \cite{riviere_brainvisa_2009}, used a skeleton representation of folds and graph-based models to directly extract the topology of cortical folds from voxel space~\cite{borne_automatic_2020}. While these methods established the foundational principles of sulcal identification, methods working in 3D volumetric space often struggle to capture the intrinsic topology of the cortex.

Consequently, the field has transitioned toward surface-based representations (triangular meshes), popularized by pipelines like FreeSurfer~\cite{fischl_cortical_1999}. Modeling the cortex as a 2-manifold is now considered essential for accurately characterizing morphology and aligning anatomy across subjects~\cite{robinson_multimodal_2018,dale_cortical_1999,fischl_cortical_1999}. This shift allows for the analysis of features, such as curvature and sulcal depth, directly on the native geometry of the cortex, a prerequisite for precise topographical labeling~\cite{hao_automatic_2020}.

The application of Geometric Deep Learning, particularly Spherical CNNs~\cite{cohen_spherical_2018} and Graph Convolutional Networks (GCNs)~\cite{monti_geometric_2017}, has shown promise in parcellating the cortical surface. However, exhaustive sulcal labeling remains an open challenge due to two critical constraints. First, state-of-the-art spherical U-Nets and similar surface-based architectures typically restrict their scope to a small subset of sulci (or only primary ones)~\cite{hao_automatic_2020,lyu_labeling_2021,parvathaneni_improving_2019}. For instance, recent work by  Hao et al.~\cite{hao_automatic_2020} focuses on 6 to 13 sulci within specific regions (like the lateral prefrontal cortex) rather than  a dense,  whole-brain labeling of 63 structures as in \cite{borne_automatic_2020}. Second, these models are data-hungry: robust performance typically requires massive cohorts (e.g., $N > 1400$ in \cite{parvathaneni_improving_2019}), making them inapplicable in expert-driven regimes where the complete annotated data is scarce ($N=62$). 
Furthermore, the recent paradigm shift toward foundation models~\cite{simeoni_dinov3_2025,kondepudi_health_2025,cox_brainsegfounder_2024} in neuroimaging have demonstrated the power of learning universal representations from massive datasets, these models currently lack native support for surface-based topologies, which limits their direct application to highly convoluted cortical surfaces.

To bridge this gap, we propose a Spherical Transfer Learning framework that enables exhaustive identification of a comprehensive set of surface-consistent sulci in a severely data-constrained regime ($N=62$). Our approach begins with a locally-optimized self-supervised pre-training on \textbf{30,000} subjects of the UK Biobank \cite{sudlow_uk_2015}. By forcing the network to focus on local surface patterns through targeted augmentations, it acquires  \virg{geometric intuition} sensitive to fine-grained anatomical nuances, such as sulcal interruptions. To transition this knowledge to the downstream task, we introduce a soft-initialized   Topological Prior Injector (TPI) that bridges the domain gap between the two-channel geometric pre-training and the three-channel semantic labeling input.
TPI allows the model to progressively integrate semantic information without suffering from catastrophic forgetting. This framework is supported by a surface-native pipeline that faithfully projects expert volumetric knowledge (as defined in \cite{borne_automatic_2020} and \cite{perrot_cortical_2011}) onto the spherical mesh, providing a unified workflow for the characterization of a comprehensive set of anatomically consistent sulci.

\section{Proposed Methodology}

\begin{figure}[t]
    \centering
    \resizebox{\textwidth}{!}{
    \begin{tikzpicture}[
        font=\sffamily\Large,
        >={Stealth[scale=1.1]},
        processNode/.style={draw, thick, fill=white, rounded corners, align=center, minimum height=1cm, inner sep=0.2cm},
        imagePlaceholder/.style={draw=gray, thick, dashed, fill=gray!10, align=center, minimum size=2.2cm, inner sep=0.1cm},
        panelLabel/.style={font=\Large\bfseries, anchor=north west},
        realImage/.style={align=center, inner sep=0pt},
        arrow/.style={->, thick},
        skipArrow/.style={->, thick, dashed, draw=gray!80}
    ]

        \begin{scope}[local bounding box=PanelA]
            \node[panelLabel, text=dataColor] (titleA) at (0, 0) {A. Data Preparation: Volumetric-to-Surface Pipeline};
            
            \node[realImage, minimum height=1.5cm, below right=0.5cm and -10cm of titleA] (imgVolume) {\includegraphics[width=5cm,  keepaspectratio]{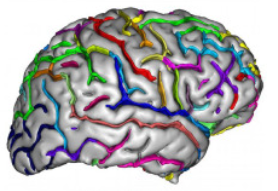}};
            \node[realImage, minimum height=1.5cm, below=0.8cm of imgVolume] (imgMesh) {\includegraphics[width=5cm,  keepaspectratio]{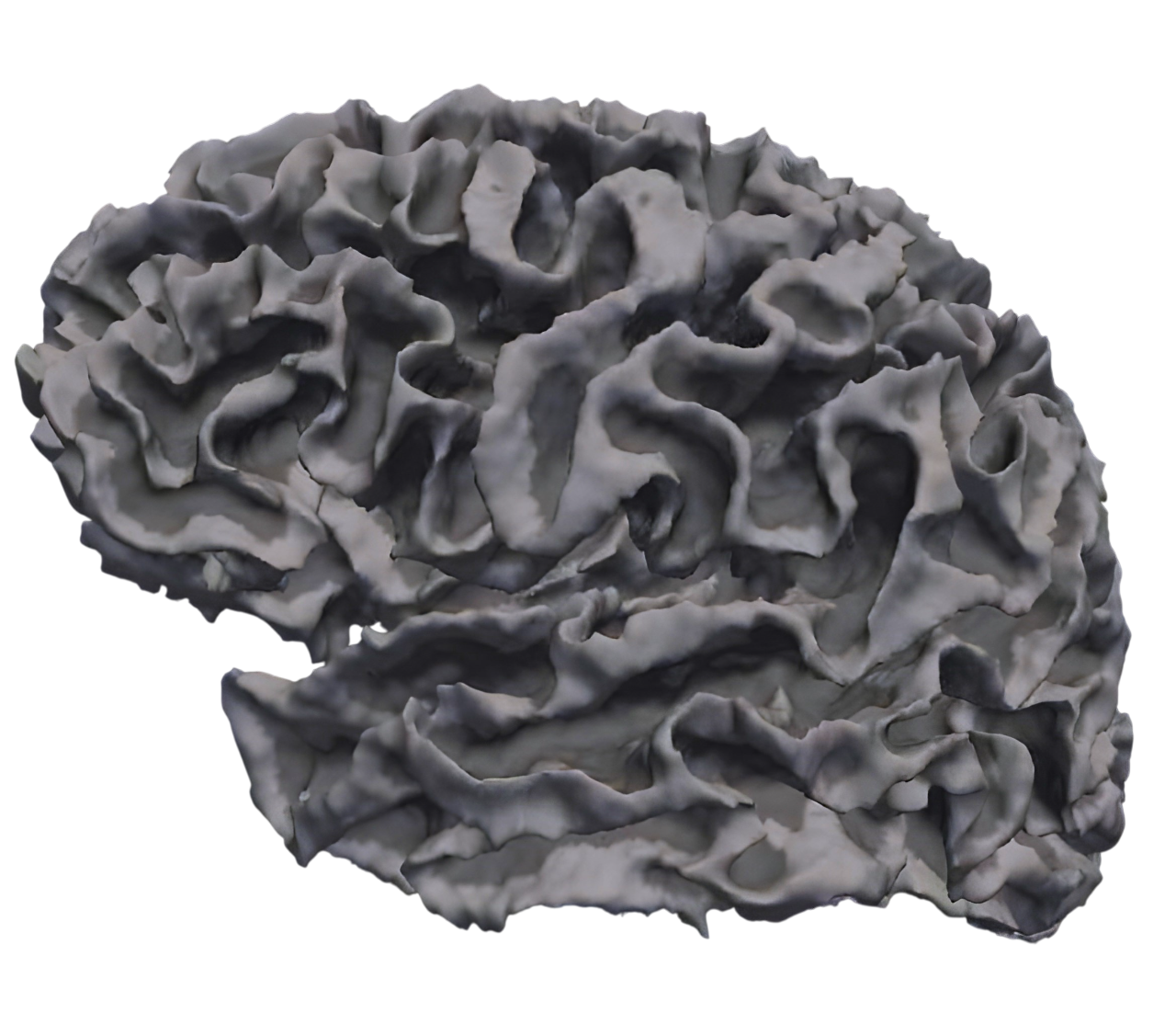}};
            
            \node[processNode, draw=dataColor, right=0.8cm of imgMesh] (trace) {TRACE Lines\\Extraction};
            
            \node[realImage, right=0.8cm of trace] (imgRawLines) {\includegraphics[width=5cm, keepaspectratio]{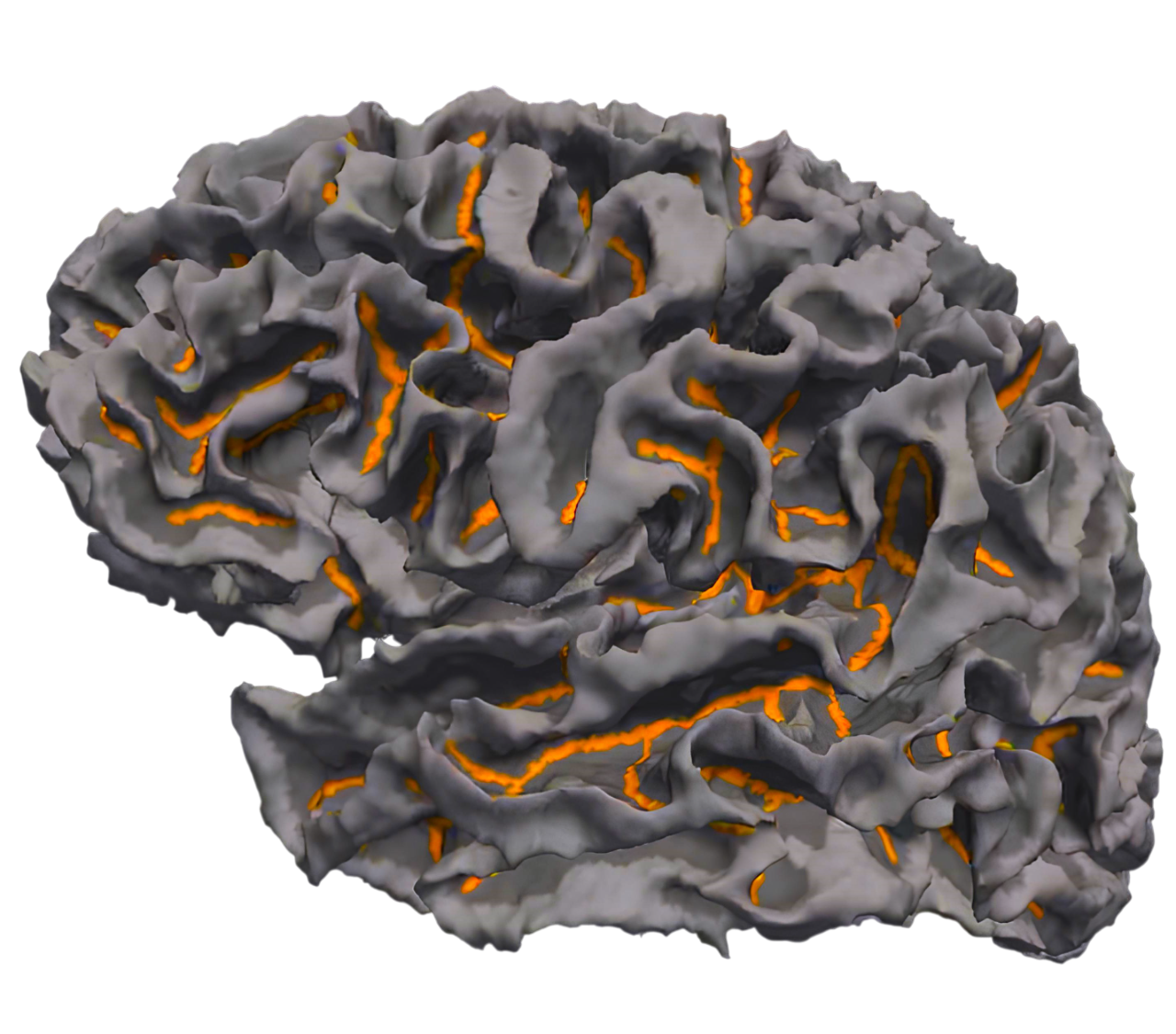}};
            
            \node[processNode, draw=dataColor, right=0.8cm of imgRawLines, yshift=1.1cm] (voronoi) {Geodesic Voronoi\\Projection};
            
            \node[realImage, right=0.8cm of voronoi] (imgColoredLines) {\includegraphics[width=6cm, keepaspectratio]{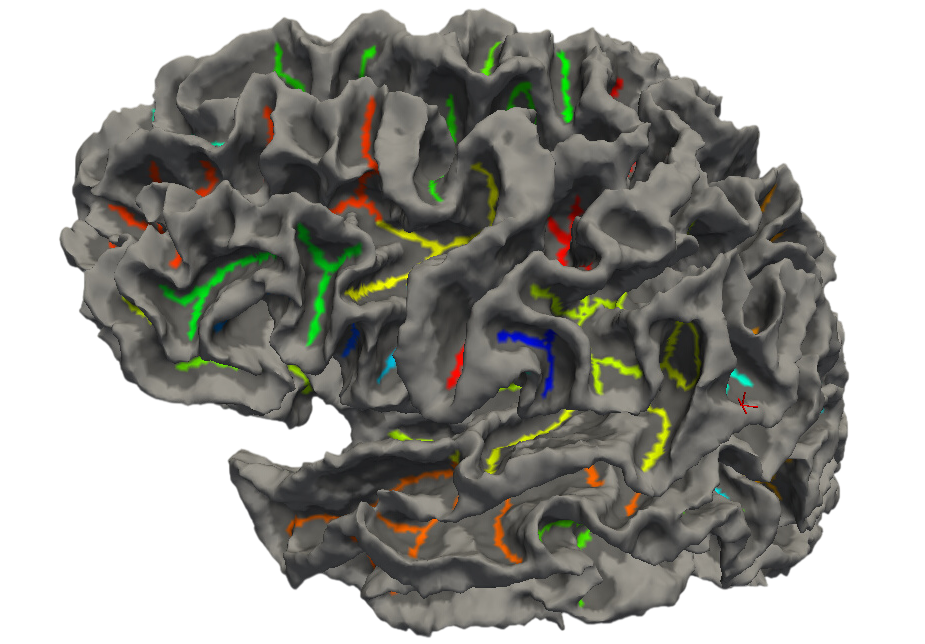}};
            
            \node[processNode, draw=dataColor, right=0.8cm of imgColoredLines] (resample) {Icosahedral Resampling\\+\\Graph Morphology};
            \node[realImage, right=0.8cm of resample] (imgSphere) {\includegraphics[width=4.2cm,  keepaspectratio]{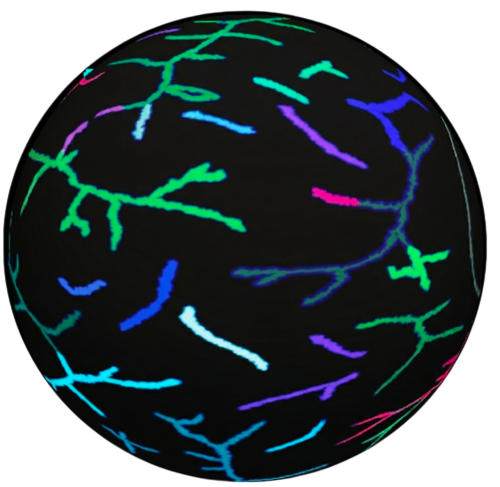}};

            \draw[arrow] (imgMesh.east) -- (trace.west);
            \draw[arrow] (trace.east) -- (imgRawLines.west);
            \draw[arrow] (imgRawLines.east) -- ++(0.3,0) |- (voronoi.200);
            \draw[arrow] (imgVolume.east) -| (voronoi.north);;
            \draw[arrow] (voronoi.east) -- (imgColoredLines.west);
            \draw[arrow] (imgColoredLines.east) -- (resample.west);
            \draw[arrow] (resample.east) -- (imgSphere.west);
        \end{scope}
        
        \begin{scope}[local bounding box=PanelB, shift={(0, -10)}]
            \node[panelLabel, text=unetColor] (titleB) at (0, 0) {B. Downstream Task: Hybrid Spherical U-Net Transfer};
            
            \node[circle, draw, thick, fill=gray!20, below right=4cm and -5cm of titleB] (inC) {$C$};
            \node[circle, draw, thick, fill=gray!20, below=0.2cm of inC] (inS) {$S$};
            \node[circle, draw, thick, dashed, fill=gray!5, below=0.2cm of inS] (inL) {$L$};
            \node[left=0.1cm of inL, font=\large, text=gray!80, align=right] {Init $\approx 0.1$};

            \node[processNode, fill=adaptColor!20, draw=adaptColor, right=1.5cm of inS, minimum height=2.5cm , minimum width=3.5cm] (adapter) {Learnable \\ TPI};
            
            \draw[arrow] (inC) -- (adapter.150);
            \draw[arrow] (inS) -- (adapter.180);
            \draw[arrow, dashed] (inL) -- (adapter.210);

            \begin{scope}[shift={(adapter.east)}, xshift=1.5cm, yshift=0cm]
                
                \filldraw[fill=sslColor!20, draw=sslColor, thick] (0, 2) -- (2.5, 1) -- (2.5, -1) -- (0, -2) -- cycle;
                \node[font=\large, text=sslColor, align=center] at (1.25, 0.35) {Encoder};
                
                \begin{scope}[shift={(1.25, -0.6)}, scale=0.8]
                    \filldraw[thick, draw=sslColor, fill=white, rounded corners=1pt] (-0.25,-0.2) rectangle (0.25, 0.2);
                    \draw[thick, draw=sslColor] (-0.15, 0.2) -- (-0.15, 0.4) arc(180:0:0.15) -- (0.15, 0.30);
                \end{scope}
                \node[font=\large, text=sslColor, align=center] at (1.25, -1.2) {Low LR};

                \node[draw=sslColor, fill=white, thick, align=center, rounded corners, font=\large, anchor=south east , minimum height=1.6cm , minimum width=7cm] (sslNote) at (0, 2.2) {Locally-Optimized SSL Prior\\(Validated on CS and PCS Task)};
                \draw[->, thick, sslColor] (sslNote.south) -- (0.15, 1.4);

                \filldraw[fill=gray!20, draw=gray, thick] (2.5, 1) rectangle (3.5, -1);
                \node[font=\large, align=center, rotate=90] at (3, 0) {Latent \\[-0.25ex] Space};

                \filldraw[fill=unetColor!20, draw=unetColor, thick] (3.5, 1) -- (6, 2) -- (6, -2) -- (3.5, -1) -- cycle;
                \node[font=\large, text=unetColor, align=center] at (4.75, 0.35) {Decoder};
                
                \begin{scope}[shift={(4.75, -0.6)}, scale=0.8]
                    \filldraw[thick, draw=unetColor, fill=white, rounded corners=1pt] (-0.25,-0.2) rectangle (0.25, 0.2);
                    \draw[thick, draw=unetColor] (-0.15, 0.2) -- (-0.15, 0.4) arc(180:0:0.15) -- (0.15, 0.50);
                \end{scope}
                \node[font=\large, text=unetColor, align=center] at (4.75, -1.2) {High LR};

                \coordinate (encTop1) at (0.3, 1.88);
                \coordinate (encTop2) at (1.1, 1.56);
                \coordinate (encTop3) at (1.9, 1.24);
                
                \coordinate (decTop3) at (4.1, 1.24);
                \coordinate (decTop2) at (4.9, 1.56);
                \coordinate (decTop1) at (5.7, 1.88);

                \draw[skipArrow] (encTop3) .. controls +(up:0.6cm) and +(up:0.6cm) .. (decTop3);
                \draw[skipArrow] (encTop2) .. controls +(up:1cm) and +(up:1cm) .. (decTop2)
                    node[midway, above, align=center, font=\large, text=black] {Local Geom. Features};
                \draw[skipArrow] (encTop1) .. controls +(up:1.6cm) and +(up:1.6cm) .. (decTop1) 
                    node[midway, above, align=center, font=\large, text=black] {\textbf{Skip Connections}};

                \coordinate (arcTop) at (3, 4.2);
                \coordinate (outNet) at (6, 0);
            \end{scope}

            \draw[arrow] (adapter.east) -- ++(1.5,0) node[midway, above, font=\large] {2 Ch.};

            \node[processNode, fill=dataColor!20, draw=dataColor, right=1cm of outNet, align=center , minimum width = 5cm] (outSoftmax) {Topological\\Post-Processing\\ \& Softmax};
           
            \node[realImage, right=1cm of outSoftmax, minimum size=1.8cm] (imgOut) {\includegraphics[width=4.2cm,  height=4.0cm]{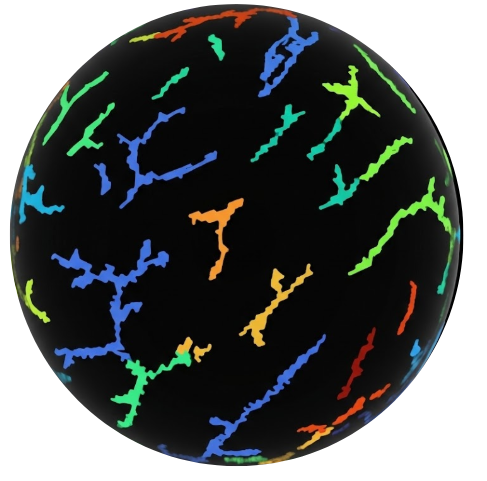}};
            
            \draw[arrow] (outNet) -- (outSoftmax);
            \draw[arrow] (outSoftmax) -- (imgOut);
        \end{scope}

    \end{tikzpicture}
    } 
    \caption{Overview of the proposed Spherical Transfer Learning framework.
    \textbf{(A)} Volumetric-to-Surface data preparation pipeline extracting 1D sulcal lines and projecting labels via Geodesic Voronoi. \textbf{(B)} Hybrid Spherical U-Net architecture. TPI  projects the 3-channel input ($C, S, L$) into the pre-trained 2-channel encoder.
    }
    \label{fig:pipeline_overview}
\end{figure}
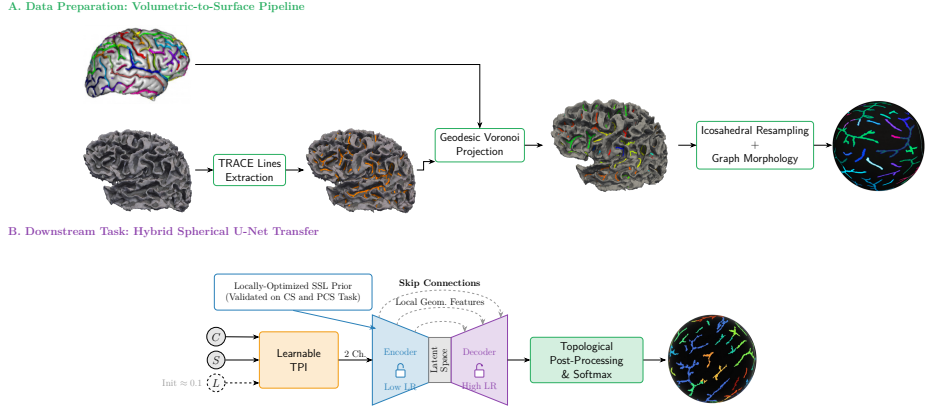

\subsection{Data Preparation and Surface-Based Pipeline}

\textbf{Dataset and Fundi Extraction: }The dataset used in this study comprises the right hemispheres of 62 subjects with expertly annotated cortical sulci, publicly available at \url{https://brainvisa.info/data/sulci_database/base_62/2019/}. We developed a  strictly surface-native pipeline extracting sulcal lines from white matter meshes via TRACE \cite{lyu_trace_2018}. Parameters were rigorously calibrated to exclusively isolate deep sulcal fundi, rejecting superficial dimples and wall ascensions.

\textbf{Volumetric-to-Surface Label Projection: }The ground-truth volumetric labels are subsequently mapped onto these 1D surface lines via a projection process based on geodesic Voronoi diagrams using a Fast Marching algorithm. To prevent aberrant semantic overlaps, such as a label improperly crossing a gyrus, this propagation is strictly constrained by the white matter mask and bounded by a maximum empirical   geodesic distance threshold of 3.5 mm.

\textbf{Spherical Input and Topological Regularization: } To provide a standardized input structure for the neural network, the native meshes from FreeSurfer \cite{fischl_cortical_1999} are parameterized onto a unit sphere and resampled onto a regular icosahedron of order 6 ($\approx 42,000$ vertices). We use the unaligned spherical parameterization to preserve local intrinsic folds, deliberately avoiding standard non-linear registration that could distort sulcal geometry. This choice increases the labeling difficulty, yet it is essential because folding patterns are not all compatible for a diffeomorphism-based alignment. 


To prevent the emergence of artificial gaps within the sulcal fundi or the loss of local structural information after interpolation, we implemented a conservative projection strategy akin to graph-based spatial max-pooling: each target icosahedral vertex aggregates the native labels within its receptive field and retains the non-zero majority value. This transfer is immediately followed by graph-based mathematical morphology operations (specifically, a spatial closing via majority voting coupled with local erosion) applied iteratively in both the native and icosahedral spaces. The anatomical viability was  validated through expert quality control on the selected 50 surface-consistent sulci.



\subsection{Locally-Optimized Spherical Pre-training (SSL)}

We employ a VGG-style spherical convolutional  encoder to align with standard state-of-the-art U-Net architectures \cite{zhao_spherical_2019} used for cortical surface segmentation, the primary focus of this work being the  adaptations made to maximize representation transferability.
The network cascades spherical convolution blocks with local pooling operations. 
Unlike standard SSL frameworks that rely on Global Average Pooling (GAP) to foster spatial invariance, we directly flatten the final convolutional output into a high-dimensional vector. Because GAP suboptimally suppresses localization  essential for  labeling, omitting it ensures a more exhaustive spatial representation .
The encoder is pre-trained on 30000 subjects of  UK Biobank using the Barlow Twins self-supervised objective\cite{zbontar_barlow_2021}. We generate augmented views using random spherical rotations with two structural occlusion strategies: random vertex dropout and contiguous patch masking. 

To verify that this globally trained latent space successfully encodes localized geometrical variations, we selected models by evaluating the frozen representations via linear probing on standard tasks (age regression, sex classification) and two topological ones:
\begin{itemize}
  
\item Central Sulcus (CS) Interruption: A rare, highly localized anatomical anomaly (< 1\%). We evaluated the model on 207 interrupted cases with 207 control cases identified using ChampollionV0 \cite{laval_towards_2024}   classifier and verified by experts.

\item Paracingulate Sulcus (PCS) Presence: A more balanced trait located in the anterior cingulate cortex \cite{cachia_shape_2014,tissier_sulcal_2018}. Labels were generated following the procedure in \cite{dufournet_self-supervised_2025}. To ensure label quality, we strictly filtered the dataset, retaining only high-confidence predictions ( $> 0.8$ for presence, $< 0.2$ for absence).

\end{itemize}

\subsection{Downstream Architecture: Transfer to the Hybrid U-Net}
The identification task requires a 3-channel input as shown in Fig. 1 (curvature $C$, sulcal depth $S$, and extracted sulcal lines $L$ as topological priors), which creates a dimensional mismatch with the 2-channel pre-trained SSL encoder (which only uses $C$ and $S$ to avoid the trivial solutions that explicit sulcal lines would provide in the topological tasks ). Naively expanding the first convolutional layer would corrupt the pre-learned filters and cause catastrophic forgetting.

To bridge this gap, we insert a \textbf{Topological Prior Injector (TPI)} directly before the encoder, implemented as a $1\times1$ convolution. Let $X \in \mathbb{R}^{V \times 3}$ be the input matrix for $V$ vertices. It projects $X$ into the 2-channel space $Y \in \mathbb{R}^{V \times 2}$ expected by the backbone:$$Y = XW + b \hspace{0.6cm} \text{and} \hspace{0.6cm} W_{init} = \begin{pmatrix} 1 & 0 \\ 0 & 1 \\ \epsilon_1 & \epsilon_2 \end{pmatrix}$$
where $W \in \mathbb{R}^{3 \times 2}$ and $b \in \mathbb{R}^2$. 
To protect the pre-trained weights from abrupt distribution shifts, the novelty lies in the soft initialization of $W$ with  $\epsilon_1, \epsilon_2 \ll 1$.
Initially, the network processes the exact geometric distribution it learned during pre-training ($Y_1 \approx C$, $Y_2 \approx S$). As fine-tuning progresses, the TPI smoothly integrates the semantic lines without abruptly shifting the input distribution.

We construct the Hybrid U-Net by grafting a randomly initialized, symmetric spherical decoder onto the pre-trained backbone. To handle small batch sizes, we used Group Normalization throughout the network. During fine-tuning, we apply differential learning rates: a low learning rate for the encoder to preserve its learned geometric manifold, and a standard rate for the TPI and decoder. 
Finally, spatial skip connections are added : they directly route the finely resolved local geometric features captured by the early SSL layers to the decoder, enforcing precise, dense labeling boundaries.

\section{Experimental Setup and Evaluation Strategy}

\subsection{Baselines and Comparative Models}
\paragraph{\textbf{Dataset and Evaluation Strategy: }}

We evaluate our framework on a cohort of 62 subjects with  surface-native sulcal labels (as detailed in 2.1). To prevent data leakage and ensure an unbiased evaluation, we implemented a dataset partition:

\textbf{-- Development Set ($N=40$):} Utilized for hyperparameter optimization via 4-fold CV. 

\textbf{-- Held-out Test Set ($N=22$):} Isolated for final  statistical testing.


\paragraph{\textbf{Baseline Configurations}} 

To isolate the specific contributions of our locally-optimized spherical SSL framework, we compare our approach against both spherical and planar baselines. 
Notably, no large-scale foundation models currently support native spherical topology. 
We utilize a gnomonic cubemap projection, mapping the sphere onto six cube faces to mitigate the polar singularities inherent to equirectangular projections. In addition to the 3D U-Net volumetric baseline from Borne et al. \cite{borne_automatic_2020} evaluated on this exact dataset, we define three surface-based baselines to isolate our approach's specific contributions: 
\begin{enumerate}
    \item Spherical U-Net "From Scratch": We trained the Spherical U-Net architecture. This quantifies the benefit of the pre-learned geometric priors .
    \item 2D U-Net (ResNet-18):  U-Net architecture with a ResNet-18 backbone.
    \item  2D Vision Foundation Models : to adapt DINOv3 \cite{simeoni_dinov3_2025} for dense sulcal labeling, we employed two decoding strategies. First, we utilize a linear probing approach with unfreezing the final two Transformer blocks.  A 1 × 1
convolution maps geometric inputs to RGB space. As a second strategy, to better capture fine-grained topology, we draw inspiration from the Dense Prediction Transformer (DPT) architecture\cite{ranftl_vision_2021}. Feature representations are extracted from layers 3, 6, 9, and 12 of the backbone network and subsequently fused using a decoder.  Implementation details can be found in our code repository.

\end{enumerate}

\subsection{Implementation Details and Hyperparameter Optimization}
During the SSL phase, models were optimized with a batch size $\in [128, 1024]$, a learning rate $\in [10^{-5}, 10^{-3}]$, and $\lambda \in [0.05, 20]$. The structural augmentations ( Sec. 2.2) were bounded as follows: Gaussian spherical rotations ($\sigma \in [2^\circ, 80^\circ]$), patch masking ($500$ to $15,000$ vertices), and vertex dropout ($0\%$ to $60\%$).

Hyperparameters were calibrated via Random Search across a 4-fold cross-validation strictly confined to the development set.  

For downstream fine-tuning, the TPI and decoder learning rates were sampled $\in [10^{-6}, 10^{-3}]$ (with the pre-trained encoder's learning rate strictly being the one-tenth of this value $\in [10^{-7}, 10^{-4}]$ (Sec. 2.3)). Weight decay was bounded $\in [0, 10^{-4}]$. The compound objective balanced a class-weighted Cross-Entropy and a Dice loss (Dice weight $\in [0, 3]$).
Spatial augmentations, applied with probability $p \in [0.3, 1.0]$, included feature noise ($\sigma \in [0, 0.4]$) and rotations ($\sigma \in [5^\circ, 90^\circ]$). Empirically, 2D projected baselines performed better without rotational augmentations; so, rotations were set to zero for these specific models. Raw predictions are refined via a  mesh-adjacency graph, where isolated connected components smaller than a threshold ($V_{min}=10$) are reassigned to the mode of their robust topological neighbors.

\section{Results}

\textbf{Evaluation metrics: }We compute the Error of Sulcal Identification (ESI), defined as a weighted sum of labeling errors across all sulci $L$: $\text{ESI} = \sum_{l \in L} w_l * \frac{FP_l + FN_l}{FP_l + FN_l + 2*TP_l}$.  ESI is intrinsically weighted by sulcal length (wl). Because length definition varies  between  3D folds and 2D lines, direct ESI comparisons are irrelevant between them.
Furthermore, anatomical analysis is conducted using the localized error rate $E_{local}(l) = \frac{FP_l + FN_l}{FP_l + FN_l + TP_l}$. It reaches 1 when a sulcus is completely hallucinated or missed (frequently for highly variable, minor sulci).

\begin{figure}[ht]
    \centering
 
    \includegraphics[width=11.5cm,height=7.5cm]{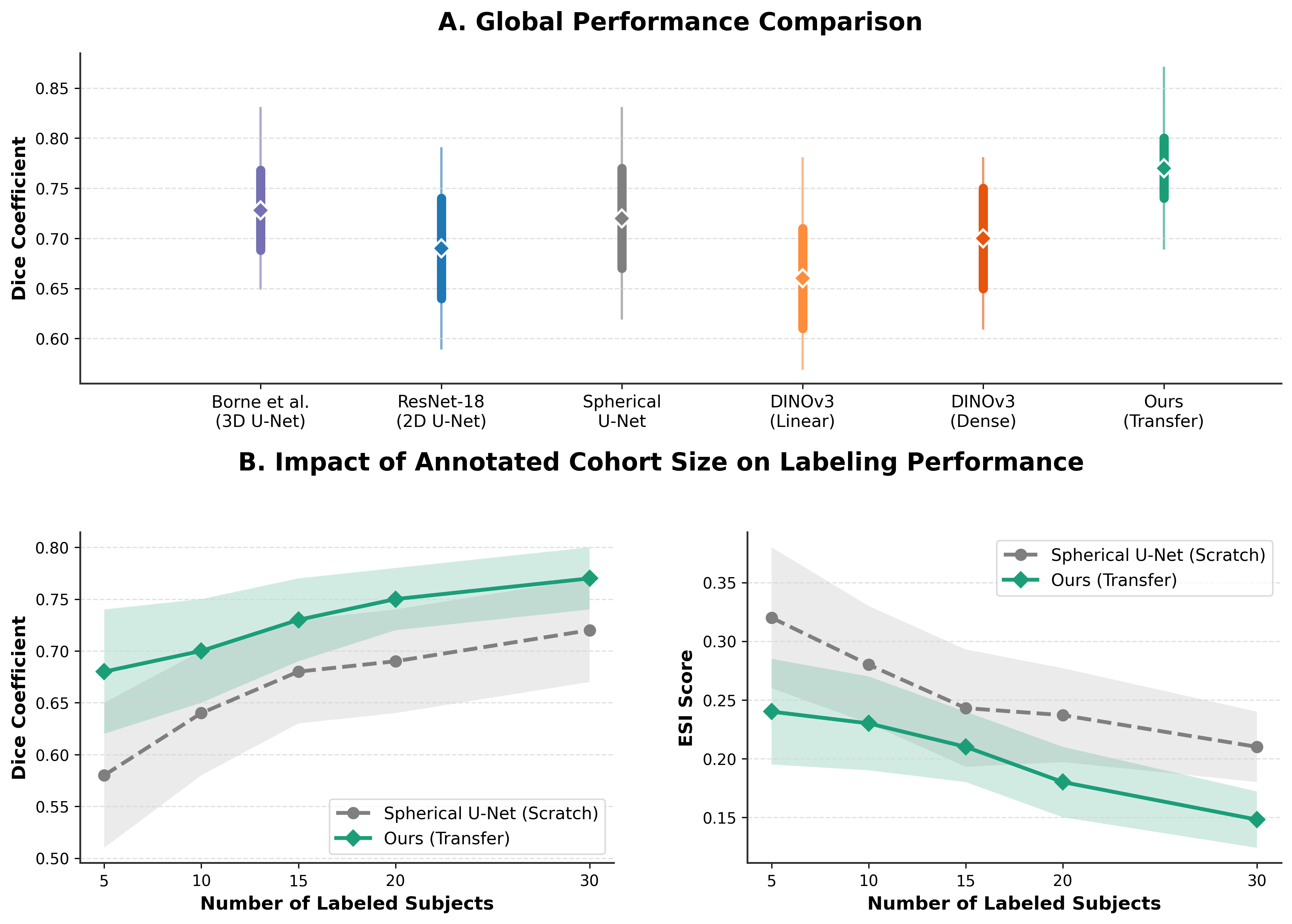}
    \caption{Quantitative evaluation of the Spherical Transfer Learning framework. 
    }
    \label{fig:performance}
   
\end{figure}

\textbf{Global Performance and Data Efficiency:}
As shown in Fig. 2A, our Spherical Transfer framework achieves the highest mean Dice ($0.77 \pm 0.03$)
on the held-out test set when trained on the maximum available split (N = 30).
Our approach visibly reduces performance variance and raises the minimum accuracy bound, demonstrating enhanced robustness to inter-subject variability. Planar projection strategies, including the vision foundation model DINOv3 (Dense: $0.70 \pm 0.05$), perform similarly to a standard 2D ResNet-18 ($0.69 \pm 0.05$). DINOv3's inability to significantly surpass from-scratch baselines highlights a structural domain gap between natural RGB images and distorted cortical topologies.

The practical value of the geometric SSL priors is most prominent in extreme low-data configurations (Fig. 2B). 
All training configurations (from 5 to 30 subjects) are strictly evaluated on the same fixed test set (N=22).
When the annotated cohort is reduced to merely 5 subjects,  the from-scratch spherical baseline exhibits severe topological degradation, with its ESI deteriorating to $0.32 \pm 0.06$. In contrast, our transferred model maintains stable identification score (ESI: $0.24 \pm 0.045$) (Wilcoxon signed-rank test, p<0.01).


\begin{table}[ht]
\centering
\caption{Ablation study evaluating the impact of pre-training objectives and transfer initialization on downstream labeling performance. 
}
\label{tab:ablation}

\begin{tabular}{l|c|c|c}{\small}

\textbf{Pre-training Strategy} & \textbf{Encoder Status} & \textbf{ Dice} ($\uparrow$) & \textbf{ESI} ($\downarrow$) \\

\hline
None ("From Scratch") & Unfrozen & $0.72 \pm 0.05$ & $0.21 \pm 0.03$ \\

\hline
Global Supervised (Age/Sex) & Frozen & $0.58 \pm 0.07$ & $0.32 \pm 0.05$ \\
Global Supervised (Age/Sex) & Unfrozen & $0.70 \pm 0.05$ & $0.22 \pm 0.04$ \\

\hline
Locally-Optimized SSL & Frozen & $0.71 \pm 0.04$ & $0.20 \pm 0.04$ \\

\textbf{Locally-Optimized SSL } & \textbf{Unfrozen} & $\mathbf{0.77 \pm 0.03}$ & $\mathbf{0.15 \pm 0.02}$ \\

\end{tabular}

\end{table}

\textbf{Ablation Study: Disentangling Pre-training Strategies: }
Table 1 isolates the specific contributions of our pre-training and transfer mechanisms. The most critical observation lies in the performance of the frozen locally-optimized encoder. Without updating any backbone weights during the downstream task, it achieves a mean Dice of $0.71 \pm 0.04$, performing on par with the fully trained from-scratch baseline ($0.72 \pm 0.05$, p>0.05). This confirms that the localized SSL pretext task successfully disentangles fine-grained cortical geometry, encoding intrinsic topological boundaries without any semantic supervision.\\
Conversely, a frozen backbone pre-trained on macroscopic targets (Age/Sex) degrades severely ($0.58 \pm 0.07$), proving that global optimization discards the high-frequency structural gradients essential for dense segmentation. While the frozen local encoder provides a robust geometric foundation, reaching peak performance (Dice: $0.77 \pm 0.03$, ESI: $0.15 \pm 0.02$) requires unfreezing the backbone (with differential learning rates) and adding our soft-initialized TPI. This full transfer framework significantly outperforms the from-scratch baseline ($p < 0.01$).

\begin{figure}[h]
    \centering
   
    \includegraphics[width=9cm]{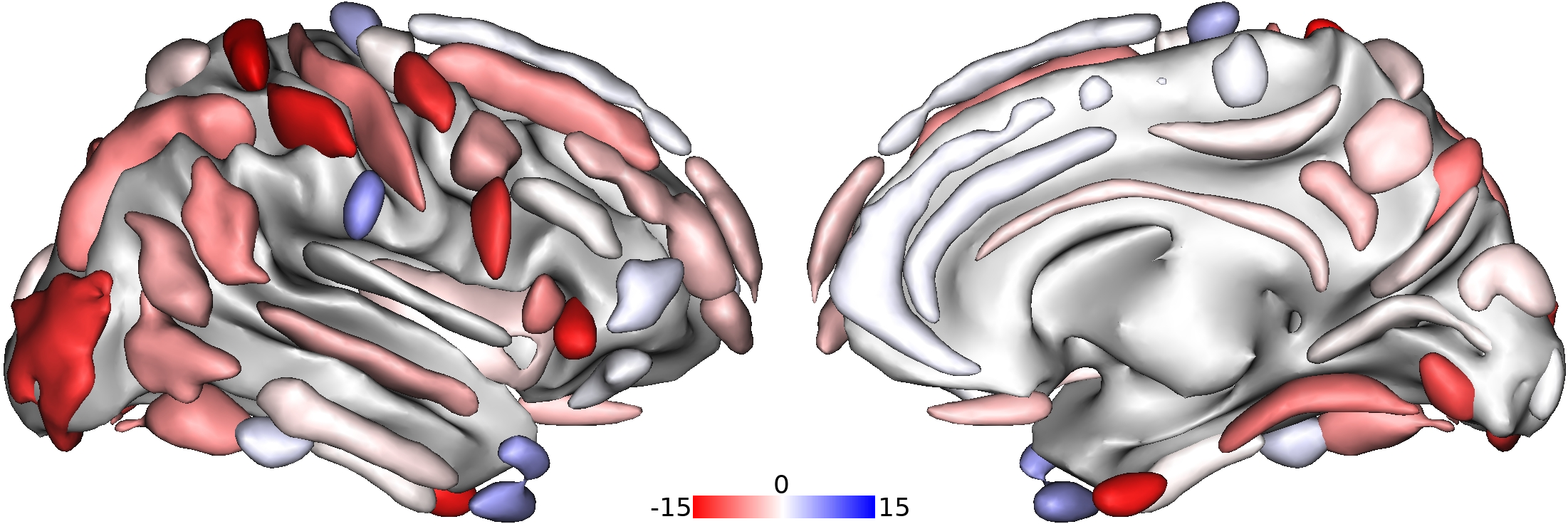}
    \caption{Surface map of local error variation ($\Delta E_{local}$) against the from-scratch model. }
    
    
    \label{fig:heatmap}
\end{figure}

\textbf{Local Error Analysis: Resolving Topological Ambiguity: }Figure 3 maps the local error variation ($\Delta E_{local}$) across the sulci, where negative values denote an error reduction  relative to the from-scratch baseline. While our transfer framework improves the delineation of certain large, prominent folds (e.g., Occipital Lobe : $-12\%$, Superior Frontal Sulcus: $-6.5\%$), the largest error reductions are concentrated on some small, highly variable structures. For instance, the Superior Postcentral ($-14.82\%$), Rhinal ($-13.94\%$), and Superior Precentral ($-12.23\%$) sulci exhibit significant performance leaps (p<0.05). By encoding intrinsic curvature and depth variations during pre-training, the network acquires the necessary topological constraints to accurately identify  ambiguous folds that are otherwise systematically smoothed over or missed in data-scarce regimes.

\section{ Conclusion}

In this work, we introduced a Spherical Transfer Learning framework to address the challenge of exhaustive cortical sulci identification in data-scarce  regimes. Our results demonstrate that decoupling geometric representation learning from semantic fine-tuning (achieved via a locally-optimized SSL prior and a soft-initialized TPI) effectively mitigates the statistical regularity bias inherent to standard supervised models. Notably, this geometric disentanglement yields the most significant identification error reductions on highly variable, tertiary sulci.

Rather than relying on architectural complexity, this study demonstrates the fundamental necessity of localized geometric priors for surface-native neuroanatomy. Future work will focus on scaling these localized pretext tasks to higher-capacity  paradigms to further advance dense cortical mapping.

\begin{credits}
\subsubsection{\ackname} This research was conducted using the \textbf{UK Biobank} resource
under application number \textbf{64984}. This project has been funded
by ANR via the Audace research program of the CEA, the IHU ICE (ANR-23-
IAHU-0010), by AXA foundation for the project premaIA, by the
"Fondation de France" for the project TSA-ideogrammes, by the PEPR
Digital Health via BHT (ANR-22-PESN-0012) and the Fondation Paralysie
Cerebrale to the ENSEMBLE project.This project was provided with computing HPC and storage resources by
\textbf{GENCI TGCC}, thanks to the grant \textbf{2024-A0170313800} on
the \textbf{Joliot Curie’s ROME} supercomputer partition.
This project was provided with computer and storage resources by
\textbf{GENCI at IDRIS} thanks to the grant \textbf{AD010316018} on the
supercomputer Jean Zay’s H100 partition.

\subsubsection{\discintname}
The authors have no competing interests to declare that are
relevant to the content of this article. 
\end{credits}

\bibliographystyle{splncs04}
\bibliography{miccai_papers}

@article{borne_automatic_2020,
	title = {Automatic labeling of cortical sulci using patch- or {CNN}-based segmentation techniques combined with bottom-up geometric constraints},
	volume = {62},
	issn = {1361-8415},
	journal = {Medical Image Analysis},
	author = {Borne, Léonie and Rivière, Denis and Mancip, Martial and Mangin, Jean-François},
	year = {2020},
	pages = {101651},
}

@article{zhao_spherical_2019,
	title = {Spherical {U}-{Net} on {Cortical} {Surfaces}: {Methods} and {Applications}},
	volume = {11492},
	issn = {1011-2499},
	shorttitle = {Spherical {U}-{Net} on {Cortical} {Surfaces}},
	urldate = {2025-09-29},
	journal = {Information processing in medical imaging : proceedings of the ... conference},
	author = {Zhao, Fenqiang and Xia, Shunren and Wu, Zhengwang and Duan, Dingna and Wang, Li and Lin, Weili and Gilmore, John H and Shen, Dinggang and Li, Gang},
	month = jun,
	year = {2019},
	pmid = {32180666},
	pmcid = {PMC7074928},
	pages = {855--866},
}

@inproceedings{cohen_spherical_2018,
	title = {Spherical {CNNs}},
	language = {en},
	urldate = {2025-09-29},
	author = {Cohen, Taco S. and Geiger, Mario and Köhler, Jonas and Welling, Max},
	month = feb,
	year = {2018},
}

@inproceedings{monti_geometric_2017,
	address = {Honolulu, HI},
	title = {Geometric {Deep} {Learning} on {Graphs} and {Manifolds} {Using} {Mixture} {Model} {CNNs}},
	isbn = {978-1-5386-0457-1},
	language = {en},
	urldate = {2025-09-29},
	booktitle = {2017 {IEEE} {Conference} on {Computer} {Vision} and {Pattern} {Recognition} ({CVPR})},
	publisher = {IEEE},
	author = {Monti, Federico and Boscaini, Davide and Masci, Jonathan and Rodola, Emanuele and Svoboda, Jan and Bronstein, Michael M.},
	month = jul,
	year = {2017},
	pages = {5425--5434},
}

@inproceedings{laval_towards_2024,
	address = {Berlin, Heidelberg},
	title = {Towards a {Foundation} {Model} for {Cortical} {Folding}},
	isbn = {978-3-031-78760-7},
	urldate = {2025-09-29},
	booktitle = {Machine {Learning} in {Clinical} {Neuroimaging}: 7th {International} {Workshop}, {MLCN} 2024, {Held} in {Conjunction} with {MICCAI} 2024, {Marrakesh}, {Morocco}, {October} 10, 2024, {Proceedings}},
	publisher = {Springer-Verlag},
	author = {Laval, Julien and Chavas, Joël and Troiani, Vanessa and Snyder, William and Patti, Marisa and Moyal, Mylène and Plaze, Marion and Cachia, Arnaud and Yi Sun, Zhong and Frouin, Vincent and Gori, Pietro and Rivière, Denis and Mangin, Jean-François},
	month = dec,
	year = {2024},
	pages = {78--88},
}

@article{lyu_trace_2018,
	title = {{TRACE}: {A} {Topological} {Graph} {Representation} for {Automatic} {Sulcal} {Curve} {Extraction}},
	volume = {37},
	issn = {1558-254X},
	shorttitle = {{TRACE}},
	number = {7},
	urldate = {2025-09-29},
	journal = {IEEE Transactions on Medical Imaging},
	author = {Lyu, Ilwoo and Kim, Sun Hyung and Woodward, Neil D. and Styner, Martin A. and Landman, Bennett A.},
	month = jul,
	year = {2018},
	pages = {1653--1663},
}

@article{fischl_cortical_1999,
	title = {Cortical {Surface}-{Based} {Analysis}: {II}: {Inflation}, {Flattening}, and a {Surface}-{Based} {Coordinate} {System}},
	volume = {9},
	issn = {1053-8119},
	shorttitle = {Cortical {Surface}-{Based} {Analysis}},
	number = {2},
	urldate = {2025-09-29},
	journal = {NeuroImage},
	author = {Fischl, Bruce and Sereno, Martin I. and Dale, Anders M.},
	month = feb,
	year = {1999},
	pages = {195--207},
}

@article{dale_cortical_1999,
	title = {Cortical {Surface}-{Based} {Analysis}: {I}. {Segmentation} and {Surface} {Reconstruction}},
	volume = {9},
	issn = {1053-8119},
	shorttitle = {Cortical {Surface}-{Based} {Analysis}},
	number = {2},
	urldate = {2025-09-29},
	journal = {NeuroImage},
	author = {Dale, Anders M. and Fischl, Bruce and Sereno, Martin I.},
	month = feb,
	year = {1999},
	pages = {179--194},
}

@article{robinson_multimodal_2018,
	title = {Multimodal surface matching with higher-order smoothness constraints},
	volume = {167},
	issn = {1095-9572},
	language = {eng},
	journal = {NeuroImage},
	author = {Robinson, Emma C. and Garcia, Kara and Glasser, Matthew F. and Chen, Zhengdao and Coalson, Timothy S. and Makropoulos, Antonios and Bozek, Jelena and Wright, Robert and Schuh, Andreas and Webster, Matthew and Hutter, Jana and Price, Anthony and Cordero Grande, Lucilio and Hughes, Emer and Tusor, Nora and Bayly, Philip V. and Van Essen, David C. and Smith, Stephen M. and Edwards, A. David and Hajnal, Joseph and Jenkinson, Mark and Glocker, Ben and Rueckert, Daniel},
	month = feb,
	year = {2018},
	pmid = {29100940},
	pmcid = {PMC5991912},
	pages = {453--465},
}

@article{lyu_labeling_2021,
	title = {Labeling lateral prefrontal sulci using spherical data augmentation and context-aware training},
	volume = {229},
	issn = {10538119},


	language = {en},
	urldate = {2026-02-17},
	journal = {NeuroImage},
	author = {Lyu, Ilwoo and Bao, Shunxing and Hao, Lingyan and Yao, Jewelia and Miller, Jacob A. and Voorhies, Willa and Taylor, Warren D. and Bunge, Silvia A. and Weiner, Kevin S. and Landman, Bennett A.},
	month = apr,
	year = {2021},
	pages = {117758},
}

@article{parvathaneni_improving_2019,
	title = {Improving human cortical sulcal curve labeling in large scale cross-sectional {MRI} using deep neural networks},
	volume = {324},
	issn = {01650270},
	language = {en},
	urldate = {2026-02-17},
	journal = {Journal of Neuroscience Methods},
	author = {Parvathaneni, Prasanna and Nath, Vishwesh and McHugo, Maureen and Huo, Yuankai and Resnick, Susan M. and Woodward, Neil D. and Landman, Bennett A. and Lyu, Ilwoo},
	month = aug,
	year = {2019},
	pages = {108311},
}

@inproceedings{hao_automatic_2020,
	address = {Iowa City, IA, USA},
	title = {Automatic {Labeling} of {Cortical} {Sulci} {Using} {Spherical} {Convolutional} {Neural} {Networks} in a {Developmental} {Cohort}},
	copyright = {https://ieeexplore.ieee.org/Xplorehelp/downloads/license-information/IEEE.html},
	isbn = {978-1-5386-9330-8},
	language = {en},
	urldate = {2026-02-17},
	booktitle = {2020 {IEEE} 17th {International} {Symposium} on {Biomedical} {Imaging} ({ISBI})},
	publisher = {IEEE},
	author = {Hao, Lingyan and Bao, Shunxing and Tang, Yucheng and Gao, Riqiang and Parvathaneni, Prasanna and Miller, Jacob A. and Voorhies, Willa and Yao, Jewelia and Bunge, Silvia A. and Weiner, Kevin S. and Landman, Bennett A. and Lyu, Ilwoo},
	month = apr,
	year = {2020},
	pages = {412--415},
}

@misc{zbontar_barlow_2021,
	title = {Barlow {Twins}: {Self}-{Supervised} {Learning} via {Redundancy} {Reduction}},
	shorttitle = {Barlow {Twins}},
	urldate = {2026-02-17},
	publisher = {arXiv},
	author = {Zbontar, Jure and Jing, Li and Misra, Ishan and LeCun, Yann and Deny, Stéphane},
	month = jun,
	year = {2021},
	note = {arXiv:2103.03230 [cs]},
}

@article{dufournet_self-supervised_2025,
	title = {A self-supervised learning framework for discovering cortical folding patterns under genetic influence: {Application} to the {Anterior} {Cingulate} {Cortex}},
	volume = {3},
	issn = {2837-6056},
	shorttitle = {A self-supervised learning framework for discovering cortical folding patterns under genetic influence},
	urldate = {2026-02-17},
	journal = {Imaging Neuroscience},
	author = {Dufournet, Antoine and Laval, Julien and Rivière, Denis and de Vareilles, Héloïse and Murray, Graham K. and Cachia, Arnaud and Chavas, Joël and Frouin, Vincent and Mangin, Jean-François},
	month = nov,
	year = {2025},
	pages = {IMAG.a.987},
}

@article{riviere_brainvisa_2009,
	series = {Organization for {Human} {Brain} {Mapping} 2009 {Annual} {Meeting}},
	title = {{BrainVISA}: an extensible software environment for sharing multimodal neuroimaging data and processing tools},
	volume = {47},
	issn = {1053-8119},
	shorttitle = {{BrainVISA}},


	urldate = {2026-02-17},
	journal = {NeuroImage},
	author = {Rivière, D and Geffroy, D and Denghien, I and Souedet, N and Cointepas, Y},
	month = jul,
	year = {2009},
	pages = {S163},
}

@misc{kondepudi_health_2025,
	title = {Health system learning achieves generalist neuroimaging models},
	urldate = {2026-02-18},
	publisher = {arXiv},
	author = {Kondepudi, Akhil and Rao, Akshay and Zhao, Chenhui and Lyu, Yiwei and Harake, Samir and Banerjee, Soumyanil and Joshi, Rushikesh and Meissner, Anna-Katharina and Hou, Renly and Jiang, Cheng and Chowdury, Asadur and Srinivasan, Ashok and Athey, Brian and Gulani, Vikas and Pandey, Aditya and Lee, Honglak and Hollon, Todd},
	month = nov,
	year = {2025},
	note = {arXiv:2511.18640 [cs]},
}

@article{sudlow_uk_2015,
	title = {{UK} biobank: an open access resource for identifying the causes of a wide range of complex diseases of middle and old age},
	volume = {12},
	issn = {1549-1676},
	shorttitle = {{UK} biobank},
	language = {eng},
	number = {3},
	journal = {PLoS medicine},
	author = {Sudlow, Cathie and Gallacher, John and Allen, Naomi and Beral, Valerie and Burton, Paul and Danesh, John and Downey, Paul and Elliott, Paul and Green, Jane and Landray, Martin and Liu, Bette and Matthews, Paul and Ong, Giok and Pell, Jill and Silman, Alan and Young, Alan and Sprosen, Tim and Peakman, Tim and Collins, Rory},
	month = mar,
	year = {2015},
	pmid = {25826379},
	pmcid = {PMC4380465},
	pages = {e1001779},
}

@misc{simeoni_dinov3_2025,
	title = {{DINOv3}},
	urldate = {2026-02-26},
	publisher = {arXiv},
	author = {Siméoni, Oriane and Vo, Huy V. and Seitzer, Maximilian and Baldassarre, Federico and Oquab, Maxime and Jose, Cijo and Khalidov, Vasil and Szafraniec, Marc and Yi, Seungeun and Ramamonjisoa, Michaël and Massa, Francisco and Haziza, Daniel and Wehrstedt, Luca and Wang, Jianyuan and Darcet, Timothée and Moutakanni, Théo and Sentana, Leonel and Roberts, Claire and Vedaldi, Andrea and Tolan, Jamie and Brandt, John and Couprie, Camille and Mairal, Julien and Jégou, Hervé and Labatut, Patrick and Bojanowski, Piotr},
	month = aug,
	year = {2025},
	note = {arXiv:2508.10104 [cs]},
}

@article{perrot_cortical_2011,
	series = {Special section on {IPMI} 2009},
	title = {Cortical sulci recognition and spatial normalization},
	volume = {15},
	issn = {1361-8415},
	number = {4},
	urldate = {2026-02-26},
	journal = {Medical Image Analysis},
	author = {Perrot, Matthieu and Rivière, Denis and Mangin, Jean-François},
	month = aug,
	year = {2011},
	pages = {529--550},
}

@misc{ranftl_vision_2021,
	title = {Vision {Transformers} for {Dense} {Prediction}},
	urldate = {2026-02-26},
	publisher = {arXiv},
	author = {Ranftl, René and Bochkovskiy, Alexey and Koltun, Vladlen},
	month = mar,
	year = {2021},
	note = {arXiv:2103.13413 [cs]},
}

@article{tissier_sulcal_2018,
	title = {Sulcal {Polymorphisms} of the {IFC} and {ACC} {Contribute} to {Inhibitory} {Control} {Variability} in {Children} and {Adults}},
	volume = {5},
	copyright = {Copyright © 2018 Tissier et al.. This is an open-access article distributed under the terms of the Creative Commons Attribution 4.0 International license, which permits unrestricted use, distribution and reproduction in any medium provided that the original work is properly attributed.},
	issn = {2373-2822},
	language = {en},
	number = {1},
	urldate = {2026-02-26},
	journal = {eNeuro},
	author = {Tissier, Cloélia and Linzarini, Adriano and Allaire-Duquette, Geneviève and Mevel, Katell and Poirel, Nicolas and Dollfus, Sonia and Etard, Olivier and Orliac, François and Peyrin, Carole and Charron, Sylvain and Raznahan, Armin and Houdé, Olivier and Borst, Grégoire and Cachia, Arnaud},
	month = jan,
	year = {2018},
	pmid = {29527565},
	note = {Publisher: Society for Neuroscience
Section: New Research},
}

@article{cachia_shape_2014,
	title = {The {Shape} of the {ACC} {Contributes} to {Cognitive} {Control} {Efficiency} in {Preschoolers}},
	volume = {26},
	issn = {0898-929X},
	number = {1},
	urldate = {2026-02-26},
	journal = {Journal of Cognitive Neuroscience},
	author = {Cachia, Arnaud and Borst, Grégoire and Vidal, Julie and Fischer, Clara and Pineau, Arlette and Mangin, Jean-François and Houdé, Olivier},
	month = jan,
	year = {2014},
	pages = {96--106},
}

@misc{cox_brainsegfounder_2024,
	title = {{BrainSegFounder}: {Towards} {3D} {Foundation} {Models} for {Neuroimage} {Segmentation}},
	shorttitle = {{BrainSegFounder}},
	urldate = {2026-02-26},
	publisher = {arXiv},
	author = {Cox, Joseph and Liu, Peng and Stolte, Skylar E. and Yang, Yunchao and Liu, Kang and See, Kyle B. and Ju, Huiwen and Fang, Ruogu},
	month = nov,
	year = {2024},
	note = {arXiv:2406.10395 [eess]},
}

\end{document}